\documentclass{article}

\usepackage{arxiv}

\usepackage[utf8]{inputenc} 
\usepackage[T1]{fontenc}    
\usepackage{hyperref}       
\usepackage{url}            
\usepackage{booktabs}       
\usepackage{amsfonts}       
\usepackage{nicefrac}       
\usepackage{microtype}      
\usepackage{lipsum}		
\usepackage{graphicx}
\usepackage{natbib}
\usepackage{doi}
\usepackage[skins, breakable]{tcolorbox}
\usepackage{listings}
\usepackage{xcolor}
\usepackage{amsmath}
\usepackage{algorithm}
\usepackage{algorithmic}

\title{Optimization through In-Context Learning and Iterative LLM Prompting for Nuclear Engineering Design Problems}

\author{ 
    \href{https://orcid.org/0000-0003-2814-1991}{\includegraphics[scale=0.06]{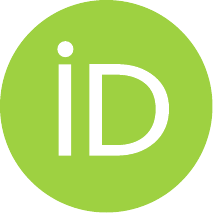}\hspace{1mm}M. Rizki Oktavian}$^{*,1,2}$, Anirudh Tunga$^{1}$, Amandeep Bakshi$^{1}$, Michael J. Mueterthies$^{1,3}$, \\ \textbf{J. Thomas Gruenwald$^{1}$, Jonathan Nistor$^{1,3}$ } \\ 
    \\
    \textsuperscript{*}{rizki@bwailabs.com} \\
    \textsuperscript{1}Blue Wave AI Labs, West Lafayette, IN, United States\\
    \textsuperscript{2}School of Nuclear Engineering, Purdue University, West Lafayette, IN, United States \\
    \textsuperscript{3}Department of Physics and Astronomy, Purdue University, West Lafayette, IN, United States
}
\date{}

\renewcommand{\headeright}{}
\renewcommand{\undertitle}{}
\renewcommand{\shorttitle}{Optimization through Iterative LLM Prompting for Nuclear Engineering Design Problems}

\hypersetup{
pdftitle={Optimization through Iterative LLM Prompting for Nuclear Engineering Design Problems},
pdfsubject={cs.LG, physics.comp-ph},
pdfauthor={M. Rizki Oktavian, Anirudh Tunga, Amandeep Bakshi, Michael J. Mueterthies,
J. Thomas Gruenwald, Jonathan Nistor},
pdfkeywords={Optimization, In-Context Learning,  Large Language Models,  Nuclear Engineering},
}

\begin{document}
\maketitle

\begin{abstract}
    The optimization of nuclear engineering designs, such as nuclear fuel assembly configurations, involves managing competing objectives like reactivity control and power distribution. This study explores the use of Optimization by Prompting, an iterative approach utilizing large language models (LLMs), to address these challenges. The method is straightforward to implement, requiring no hyperparameter tuning or complex mathematical formulations. Optimization problems can be described in plain English, with only an evaluator and parsing script needed for execution.  The in-context learning capabilities of LLMs enable them to understand problem nuances, therefore they have the potential to surpass traditional metaheuristic optimization methods. This study demonstrates the application of LLMs as optimizers to Boiling Water Reactor (BWR) fuel lattice design, showing the capability of commercial LLMs to achieve superior optimization results compared to traditional methods.
\end{abstract}

\keywords{Optimization \and In-Context Learning  \and  Large Language Models \and  Nuclear Engineering}

\begin{figure*}[h!]
    \centering
    \includegraphics[width=0.75\linewidth]{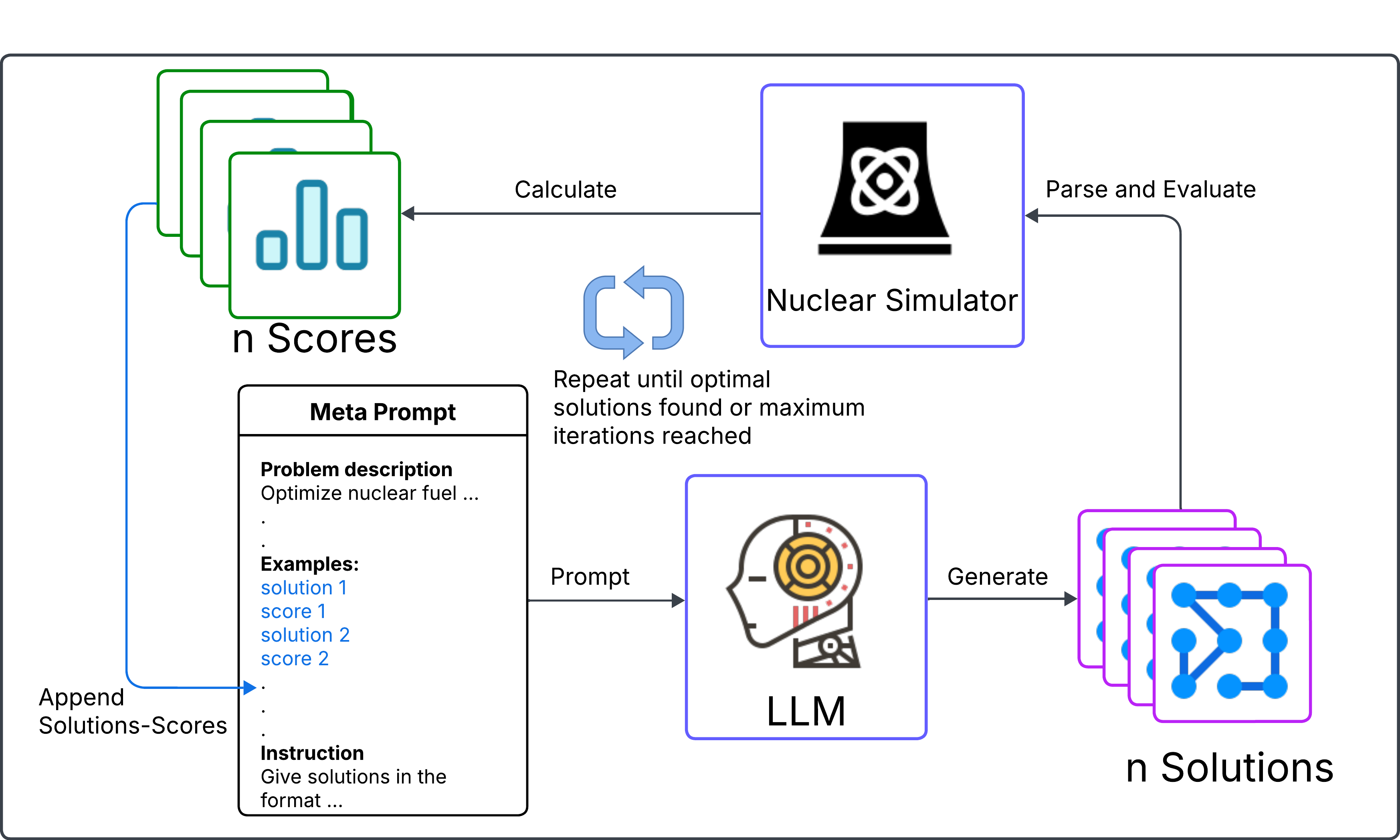}
    \text{Iterative Optimization Framework Using LLM and Nuclear Simulation}
    \label{fig:abstract}    
\end{figure*}

\section{Introduction}

The application of Large Language Models (LLMs) has mostly centered on text-based tasks such as natural language understanding and content generation \citep{minaee2024largelanguagemodelssurvey}. However, this work introduces a novel approach by utilizing LLMs for numerical optimization in engineering contexts. By integrating the logical reasoning capabilities of LLMs with high-fidelity physics-based simulations, this method bridges the gap between language-based reasoning and complex numerical problem-solving. This method allows LLMs to adjust and improve solutions over time, making them useful for optimizing engineering designs.

The use of LLMs in nuclear engineering consists of several applications. For example, NukeLM and NukeBERT have been fine-tuned on nuclear-specific datasets, enhancing performance in tasks such as classifying research and answering domain-specific questions \citep{burke2021, jain2020}. Additionally, LLMs have been applied to assist fusion energy research by processing large datasets for informed experimental decisions \citep{mehta2023}. These studies highlight how LLMs can optimize nuclear engineering processes, improve decision-making, and enhance safety analysis.

In the field of optimization, traditional optimization approaches for nuclear engineering problems typically rely on gradient-based numerical methods that require precise mathematical formalization of design objectives. Recent advances in LLM capabilities enable a new optimization approach through Optimization by Prompting (OPRO), first demonstrated by Yang et al. in their work "Large Language Models as Optimizers"\citep{yang2023}. This natural language-based optimization framework uses LLMs' reasoning abilities to improve solutions through prompt engineering iteratively.

The OPRO approach introduces three key innovations for optimization techniques: First, it replaces mathematical formulations with natural language problem descriptions, therefore enable domain experts to articulate constraints without formal optimization training. Second, it leverages LLMs' pattern recognition capabilities to identify improvement directions from historical solution-score pairs. Third, the approach provides inherent explainability through the model's chain-of-thought reasoning process\citep{wei2022}. These characteristics make OPRO particularly suitable for nuclear engineering applications where problem formalization challenges often hinder conventional optimization methods.

At the core of OPRO lies its meta-prompt architecture, which combines three elements: 1) Natural language descriptions of the optimization task, 2) Historical solutions with performance scores, and 3) Exemplary problem instances. Through iterative refinement, the LLM generates candidate solutions that are evaluated and reintegrated into subsequent prompts. This process mirrors human expert reasoning patterns while surpassing manual optimization in both speed and solution quality\citep{wang2022}.

Our application of OPRO to Boiling Water Reactor (BWR) fuel lattice optimization demonstrates the method's potential for nuclear design challenges. Focusing on the GE-14-like assembly configuration \citep{Cumberland2021, Oktavian2023TransitionPARCS/Polaris, Mertyurek2023}, we adapt the original OPRO framework to address the multi-objective optimization problem of targeting certain reactivity while minimizing power peaking factors. The LLM's ability to interpret neutronic constraints through natural language descriptions proves particularly valuable in navigating the complex parameter space.

This work extends the original OPRO methodology in three significant directions: First, we develop domain-specific prompt templates for nuclear engineering applications. Second, we introduce hybrid evaluation functions combining LLM reasoning with nuclear lattice simulations in Casmo-5 \citep{CASMO5}. Third, we demonstrate OPRO's capability to handle problem-specific constraints through prompt engineering.

\section{Optimization by Prompting (OPRO)}

Optimization by Prompting represents a new approach in numerical optimization by utilizing LLMs as meta-optimizers through natural language instructions \citep{yang2023}. Instead of relying on strict mathematical formulas, it uses natural language instructions to guide the optimization process. Notably, OPRO has demonstrated significant improvements in prompt engineering tasks, achieving up to a 50\% increase in performance compared to human-designed prompts on complex reasoning benchmarks \citep{wei2022}.

Within this framework, the optimization problem is defined over a solution space $\mathcal{S}$. It is mathematically formalized as 
\begin{equation}
    c^* = \underset{c \in \mathcal{C}}{\text{argmax}} \ \mathbb{E}_{x \sim \mathcal{D}}[\mathcal{M}(c \oplus x)],
\end{equation}
where $c$ denotes a solution, $\mathcal{D}$ is the data distribution, $\mathcal{M}$ is the LLM scoring function, and $\oplus$ indicates the concatenation operator \citep{yang2023}. The approach employs concepts such as natural language gradient descent. The process is realized through iterative prompt refinement guided by LLM-generated candidate solutions. It relies on an underlying loss landscape created from pairs of solutions and their scores.

The practical implementation of OPRO is included in an iterative refinement process that integrates a meta-prompt architecture containing a natural language problem description, historical (solution, score) pairs, optimization trajectory analysis, and candidate generation instructions. This process is formalized in the algorithm below.

\begin{algorithm}[H]
\caption{Iterative Refinement Process in OPRO}
\begin{algorithmic}[1]
\STATE \textbf{Input:} Initial meta-prompt $\mathcal{P}_0$, convergence threshold $\epsilon$
\STATE \textbf{Initialize:} $t \gets 0$
\REPEAT
    \STATE Generate candidate solutions $C_t = \{c_t^1, c_t^2, \ldots, c_t^k\}$ using the LLM optimizer
    \FOR{each candidate $c_t^i \in C_t$}
        \STATE Evaluate score $s_t^i = \mathcal{M}(c_t^i)$
    \ENDFOR
    \STATE Update meta-prompt: $\mathcal{P}_{t+1} \gets \mathcal{P}_t \cup \{(c_t^i, s_t^i) \mid \forall i\}$
    \STATE $t \gets t+1$
\UNTIL{$\max_{i}\{s_t^i\} - \max_{i}\{s_{t-1}^i\} < \epsilon$}
\STATE \textbf{Output:} Optimized solution $c^*$
\end{algorithmic}
\end{algorithm}

\section{Test Problem: BWR Fuel Lattice Optimization }

\begin{figure}[h!]
    \centering
    \includegraphics[width=0.4\linewidth]{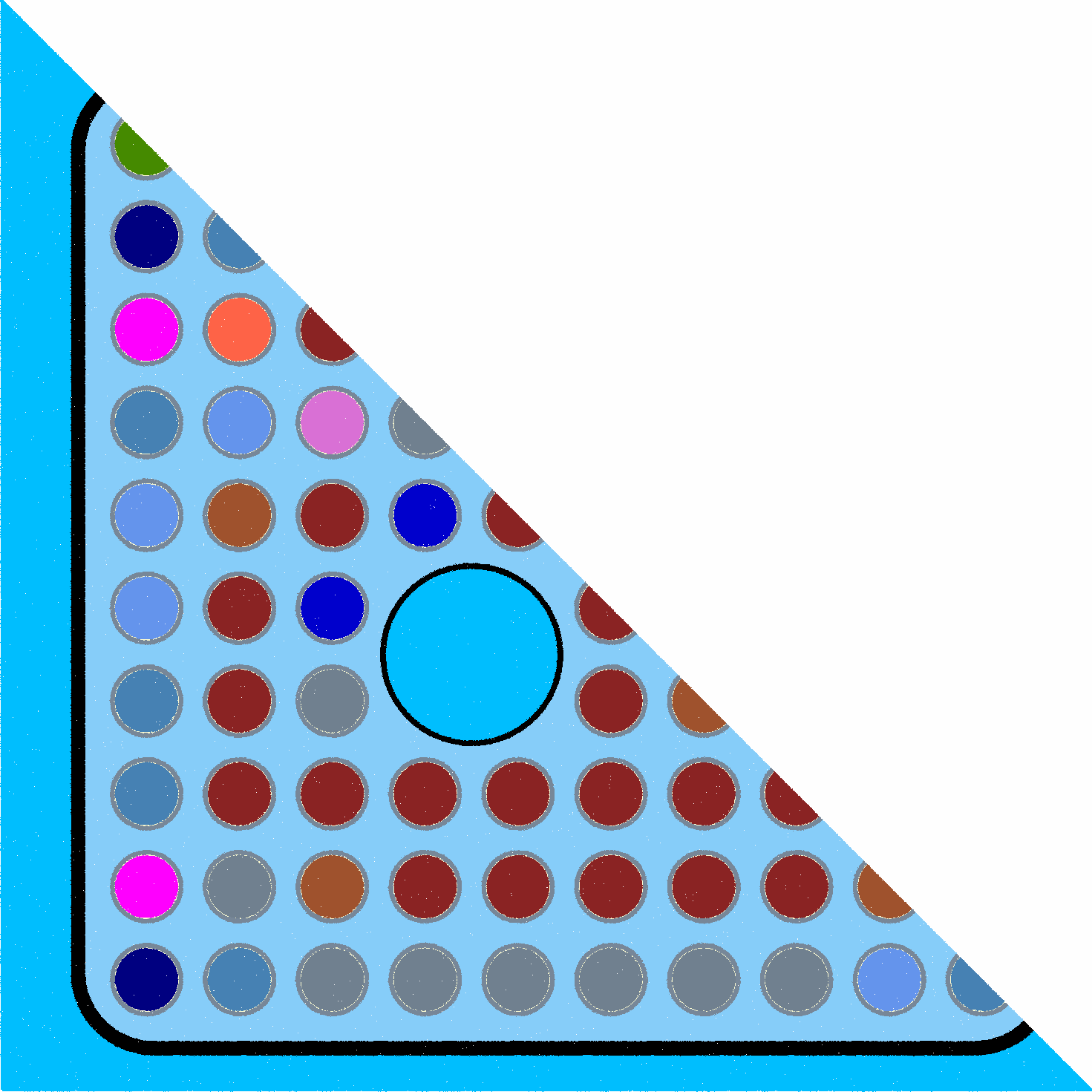}
    \caption{Half symmetry of GE-14 Dominant (DOM) fuel lattice. The image is generated by the authors using Casmo-5 software.}
    \label{fig:GE-14}
\end{figure}

This test problem focuses on optimizing the fuel lattice design for a BWR based on the GE-14 assembly configuration \cite{Cumberland2021}. The primary objective is to optimize the uranium enrichment levels of 11 fuel pins and the gadolinium (Gd) content of 4 burnable poison pins. The illustration of the BWR lattice design is presented in Figure \ref{fig:GE-14}.

The problem is defined by several key constraints and objectives. The target beginning-of-cycle (BOC) infinite multiplication factor (\(k_{\text{inf}}\)) is approximately 1.05, while the BOC power peaking factor (PPF) must remain below 1.33. The PPF is defined as the ratio of maximum power to average power within the fuel lattice. Additionally, the maximum allowable uranium-235 enrichment is 5.0\%, with increments of 0.1\%, and the gadolinium content is limited to a maximum of 10\%, adjustable in increments of 1\%.

The objective function for this optimization problem is expressed as:
\begin{equation}
    f_{\text{ob}} = 100 - w_1 \left| k_{\text{inf}} - k_{\text{inf}}^{\text{targ}} \right| - w_2 \max(0, PPF - PPF^{\text{targ}})
\end{equation}

where \(w_1\) and \(w_2\) are weighting factors that penalize deviations from the target \(k_{\text{inf}}\) and PPF values. This function ensures that solutions are rewarded for meeting both reactivity and power-peaking constraints while penalizing those that violate them.

The optimization strategy involves generating candidate solutions for pin enrichment and gadolinium distribution. These candidates are evaluated using CASMO-5, a high-fidelity neutronics simulation tool, to compute \(k_{\text{inf}}\) and PPF values.

As a note, the test problem is a simplified version of the actual fuel lattice optimization process by simulating just a single-state BOC case for every trial. While it is possible to run a full depletion case with varying states using Casmo-5, it will only scale up the parameters in the objective function. For the preliminary demonstration purpose of this approach, this simplified test problem should suffice.

\section{Approaches}

The optimization process combines the capabilities of LLM and CASMO-5 simulations to solve the BWR GE-14 lattice optimization problem. The approach utilizes the iterative refinement of solutions through meta-prompting and numerical evaluation, ensuring that generated designs meet the desired performance criteria.

The workflow begins with an initial set of solutions generated using CASMO-5. These solutions are included into the meta-prompt provided to the LLM, which generates candidate solutions for fuel pin enrichment and gadolinium distribution. The generated solutions are then evaluated in CASMO-5 simulations to compute \(k_{\text{inf}}\) and PPF. The results are used to calculate the objective function, which guides the iterative optimization process. The meta-prompt is updated with solution-score pairs after each iteration, enabling the LLM to learn from previous results and improve subsequent candidate generations. This process continues until the convergence criteria are met or performance improvements plateau.

The optimization strategy includes two prompting techniques for comparison:
\begin{itemize}
    \item \textbf{Detailed Context Prompting:} This method provides the LLM with a comprehensive description of the problem, including all relevant constraints, historical solutions, and evaluation metrics.
    \item \textbf{No Context Prompting:} In this approach, only minimal information about the problem is provided to the LLM, relying on its inherent reasoning capabilities to generate solutions.
\end{itemize}

To understand the difference between the two prompting strategies, Appendix  \ref{app:meta-promts} and Appendix \ref{app:responses} provide the textual contents of each meta-prompt as well as some sample responses from the LLM. To evaluate the effectiveness of these approaches, three variants of \textbf{Google Gemini 2.0} models\citep{geminiteam2024geminifamilyhighlycapable} are used:
\begin{itemize}
    \item \textbf{Flash-Lite:} A smaller, faster version of Gemini Flash, optimized for efficiency.
    \item \textbf{Flash:} A general-purpose model designed for a wide range of tasks.
    \item \textbf{Flash-Thinking:} A reasoning-focused model tailored for complex problem-solving.
\end{itemize}

For benchmarking purposes, a genetic algorithm is employed as a reference optimization method. This comparison allows for an assessment of how well LLM-based optimization performs relative to traditional evolutionary algorithms in terms of convergence speed, solution quality, and computational efficiency.

\begin{figure}[h!]
    \centering
    \includegraphics[width=0.75\linewidth]{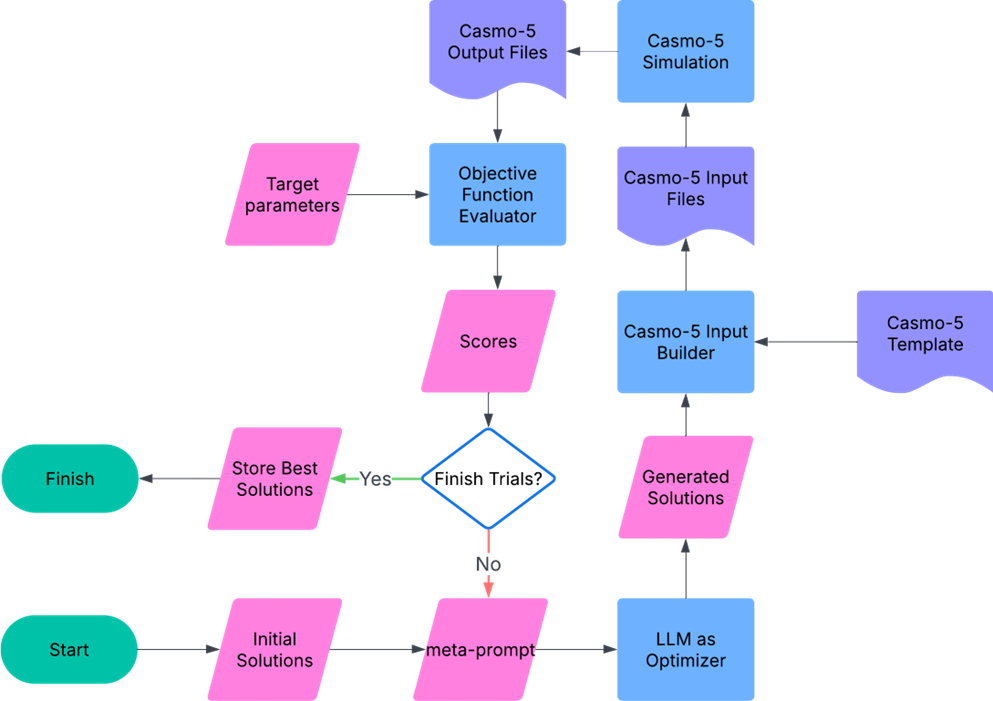}
    \caption{Optimization by LLM prompting flowchart with Casmo-5 evaluations. The image is generated by authors using Lucid App.}
    \label{fig:OPRO_Casmo5}
\end{figure}

The diagram in Figure~\ref{fig:OPRO_Casmo5} illustrates the overall workflow. The process begins with initial solutions fed into a meta-prompt for LLM-based generation. The generated solutions are then passed through CASMO-5 simulations to evaluate their performance against target parameters. Scores are calculated based on the objective function, and if trials are not complete (based on the maximum score achieved or maximum number of steps/iterations), updated solution-score pairs are fed back into the meta-prompt for further refinement. The loop continues until optimal solutions are identified or all trials are completed.

\section{Results}

The results of the optimization trials demonstrate the performance of different models and prompting techniques in achieving the highest scores and minimizing the number of steps required for convergence. The evaluation compares three variants of the Gemini 2.0 model (Flash Lite, Flash, and Flash Thinking) under two prompting strategies (detailed context meta-prompt and no context meta-prompt) against a baseline genetic algorithm.

Table~\ref{tab:best_scores} summarizes the average best scores achieved across 10 optimization trials. The Gemini 2.0 Flash Thinking model consistently achieves the highest score of \(100.00 \pm 0.00\) when using the detailed context meta-prompt, outperforming all other approaches, including the genetic algorithm (\(99.55 \pm 1.34\)). Notably, even with no context prompting, Flash Thinking achieves near-optimal performance (\(99.70 \pm 0.89\)), indicating its robustness in reasoning-based tasks. 

\begin{table}[h!]
    \caption{Best Scores for 10 Optimization Trials}
    \centering
    \begin{tabular}{|l|c|c|}
        \hline
        \textbf{Model} & \textbf{Detailed Context Meta-prompt} & \textbf{No Context Meta-prompt} \\
        \hline
        Gemini 2.0 Flash Lite & $91.55 \pm 6.41$ & $96.74 \pm 4.60$ \\\hline
        Gemini 2.0 Flash & $97.20 \pm 8.40$ & $96.43 \pm 6.13$ \\\hline
        Gemini 2.0 Flash Thinking & $\mathbf{100.00 \pm 0.00}$ & $99.70 \pm 0.89$ \\\hline
        Genetic Algorithm & \multicolumn{2}{c|}{$99.55 \pm 1.34$} \\\hline
    \end{tabular}
    \label{tab:best_scores}
\end{table}

Interestingly, smaller models like Flash Lite performed better with no context prompting (\(96.74 \pm 4.60\)) compared to detailed context prompting (\(91.55 \pm 6.41\)). This suggests that smaller LLMs may struggle with processing extensive contextual information due to their limited attention spans but can still generate competitive solutions when given simpler instructions.

The number of steps required for convergence, as shown in Table~\ref{tab:steps}, reveals that Flash Thinking achieved optimal solutions faster with no context prompting (\(12.30 \pm 17.06\)) than with detailed context prompting (\(19.10 \pm 10.13\)). However, detailed context prompting ensured consistent attainment of the maximum score across trials, albeit requiring more iterations due to the model's tendency to "overthink" during optimization.

\begin{table}[h!]
    \caption{Steps to Achieve Best Score for 10 Optimization Trials}

    \centering
    \begin{tabular}{|l|c|c|}
        \hline
        \textbf{Model} & \textbf{Detailed Context Meta-prompt} & \textbf{No Context Meta-prompt} \\
        \hline
        Gemini 2.0 Flash Lite & $40.90 \pm 18.21$ & $25.40 \pm 20.50$ \\ \hline
        Gemini 2.0 Flash & $12.40 \pm 13.54$ & $26.60 \pm 20.04$ \\ \hline
        Gemini 2.0 Flash Thinking & $19.10 \pm 10.13$ & $\mathbf{12.30 \pm 17.06}$ \\ \hline
         Genetic Algorithm & \multicolumn{2}{c|}{$23.20 \pm 14.18$}\\ 
        \hline
    \end{tabular}
    \label{tab:steps}
\end{table}

\begin{figure}[h!]
    \centering
    \includegraphics[width=1\linewidth]{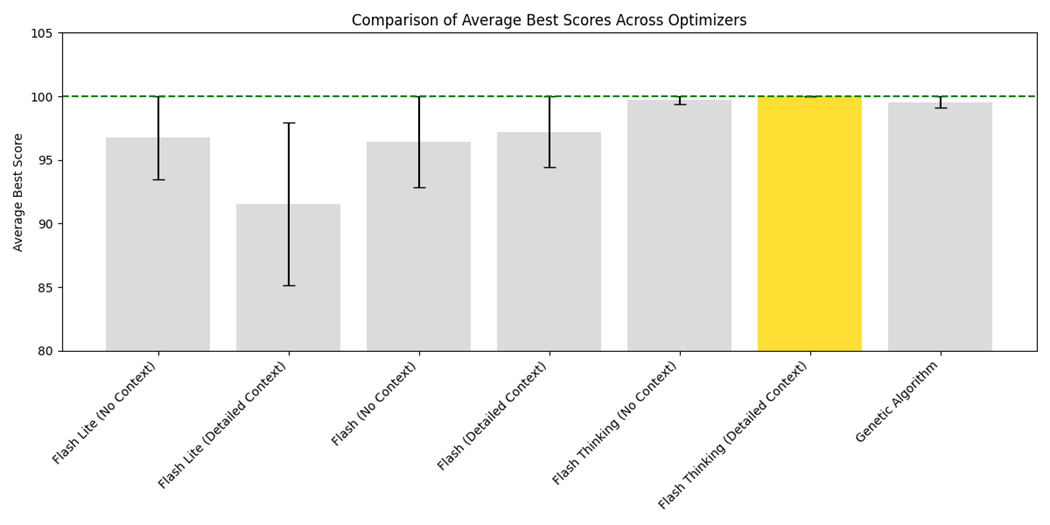}
    \caption{Comparison of Average Best Scores Across Optimizers for Different Prompting Strategies. The yellow bar highlights the optimal approach.}
    \label{fig:score_comparison}
\end{figure}

\begin{figure}[h!]
    \centering
    \includegraphics[width=1\linewidth]{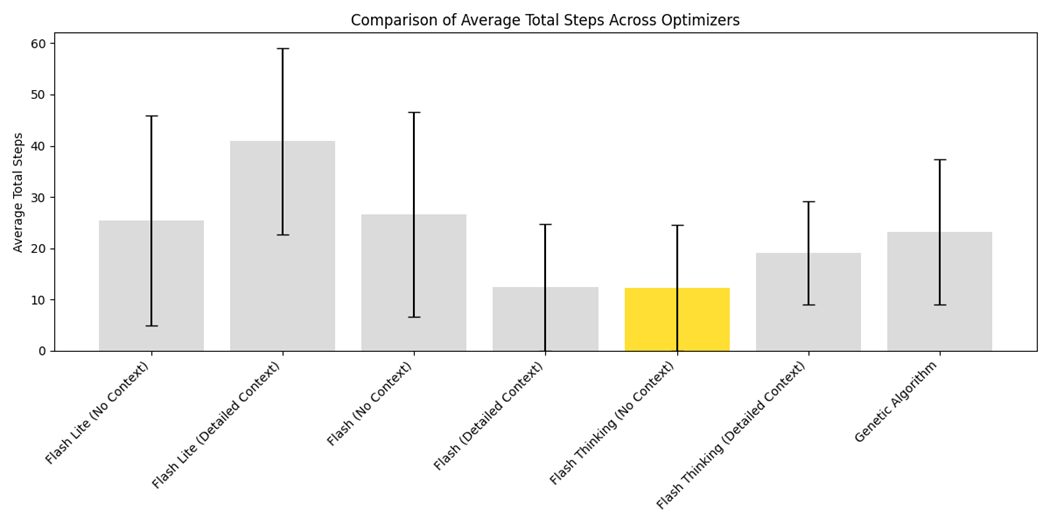}
    \caption{Average Steps Required for Convergence Across Optimizers and Prompting Strategies. The yellow bar highlights the optimal approach.}
    \label{fig:steps_comparison}
\end{figure}

Figure~\ref{fig:score_comparison} shows a comparison of average best scores across optimizers, highlighting that Flash Thinking with detailed context achieves the maximum score consistently across all trials, while no context prompting slightly reduces stability but improves speed for certain models. Figure~\ref{fig:steps_comparison} illustrates the average total steps required for convergence across optimizers, where Flash Thinking with no context prompting demonstrates superior efficiency compared to other methods.

\begin{figure}[h!]
    \centering
    \includegraphics[width=0.8\linewidth]{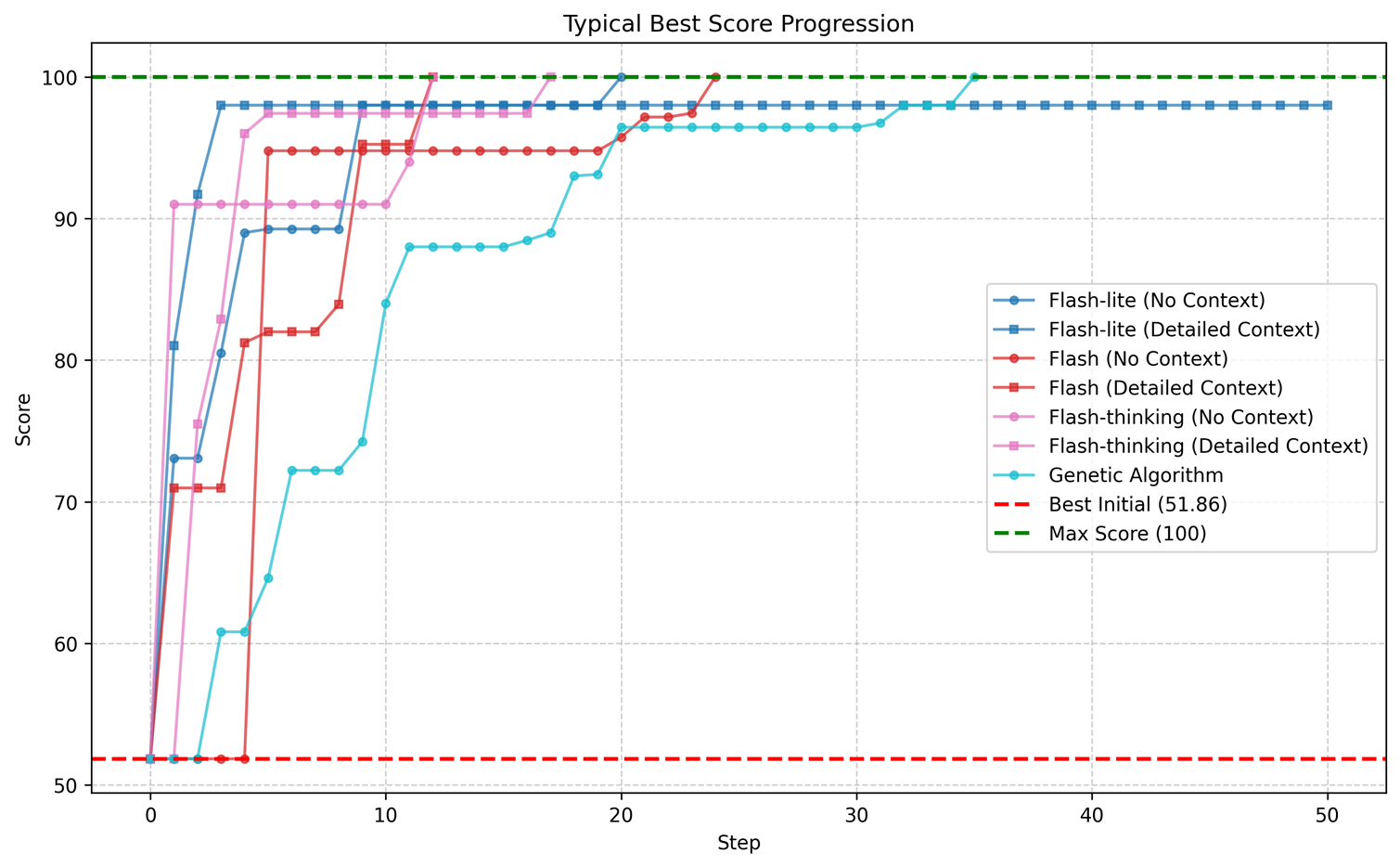}
    \caption{Progression of Best Scores Over Optimization Steps for Different Models and Methods.}
    \label{fig:progression}
\end{figure}

Additionally, Figure~\ref{fig:progression} presents the progression of best scores over optimization steps for different models and methods, showing that LLM-based approaches tend to exhibit larger oscillations in early iterations compared to the genetic algorithm, which balances exploration and exploitation more consistently.

The presented results also show that crafting effective meta-prompts plays a critical role in enhancing LLM performance. Detailed prompts help larger models interpret optimization constraints more effectively, leading to superior results. On the other hand, simpler prompts are better suited for smaller models or tasks requiring faster convergence without sacrificing too much accuracy.

Several important limitations must be considered when applying these techniques to more complex optimization problems. Scaling these approaches to problems with higher-dimensional inputs presents significant challenges. As the number of parameters increases, the process of converting lists of numbers into text-based prompts for LLMs becomes increasingly difficult. The textual representation of high-dimensional data may exceed the context window limitations of even the most advanced LLMs. 

Furthermore, as the complexity of instructions increases with higher-dimensional problems, LLMs become more prone to hallucination. When presented with complex constraints, objective functions, and an excessive number of parameters, models may struggle to generate solutions based on the requested formats. This can result in parsing errors when translating LLM outputs into numerical solutions. Even when the parsing succeeds, the model may still violate the given constraints and rules, resulting in unacceptable solutions.

\section{Conclusion}

This study highlights the simplicity and effectiveness of novel optimization using iterative LLM prompting for complex engineering tasks like BWR fuel lattice design. OPRO requires no hyperparameter tuning and allows problems to be described in plain English, with only an evaluator and a parsing script needed for implementation. LLMs’ ability to understand nuanced problem constraints gives them the potential to outperform traditional metaheuristic methods, particularly when using larger models with longer attention spans. Reasoning-focused models are recommended for consistent results, and crafting effective meta-prompts helps LLMs better interpret context. Despite these advantages, practitioners should be mindful of the scalability challenges when applying these techniques to higher-dimensional problems, where the conversion of complex parameter spaces into textual representations may exceed the prompt context window length. Additionally, as problem complexity increases, there is an elevated risk of model hallucination that could lead to suboptimal solutions.  Future advancements could benefit from LLMs specialized in nuclear engineering knowledge, further enhancing their applicability in this domain.

\section*{Acknowledgements}

This research made use of the resources of the High-Performance Computing Center at Idaho National Laboratory, which is supported by the Office of Nuclear Energy of the U.S. Department of Energy and the Nuclear Science User Facilities under Contract No. DE-AC07-05ID14517. The authors would like to thank Charles F. Karlson for his work on building Casmo-5 models for the GE-14 fuel lattice from the SCALE models.

\bibliographystyle{unsrtnat}
\bibliography{references}

\newpage

\appendix

\section{Meta-prompts}
\label{app:meta-promts}

\subsection{No Context Meta-prompt}

\begin{tcolorbox}
Your task is to generate solutions from an optimization problem.
Below are examples of solution-score pairs:

\begin{lstlisting}[basicstyle=\small\ttfamily\color{blue}]
<sol> 1.4,2.2,2.6,4.2,5.0,4.7,3.7,4.1,8.0,4.9,7.0,5.0,6.0,5.0,8.0 <\sol>
score: 44.08

<sol> 1.8,2.4,2.9,4.0,5.0,4.7,3.4,3.9,5.0,4.6,5.0,5.0,8.0,5.0,7.0 <\sol>
score: 66.6   

<sol> 1.8,2.5,2.7,4.2,5.0,4.9,3.7,3.8,8.0,4.7,5.0,5.0,8.0,4.8,9.0 <\sol>
score: 51.86

\end{lstlisting}

Generate exactly \textcolor{red}{\{batch\_size\}} new solutions that:\\
- Are distinct from all previous solutions.\\
- Have a higher score than the highest provided.\\

The solutions should start with \texttt{<sol>} and end with \texttt{<\textbackslash sol>} with a comma between parameters.     
\end{tcolorbox}

The text in \textcolor{blue}{blue} is the solution-score pairs that are added every iteration.
\textcolor{red}{\{batch\_size\}} is to generate multiple solutions for every LLM response, currently set to 5 for this work.

\subsection{Detailed Context Meta-prompt}

\begin{tcolorbox}[breakable]
You are an optimization agent and an expert in nuclear reactor design.  \\
Your task is to generate a 10×10 GE-14 fuel lattice design that satisfies the following conditions:  
\\
- Fuel Enrichment (FUE\#\_enr): Maximum 5.0\% with increments of 0.1\%.  \\
- Gadolinia Content (FUE\#\_gads): Maximum 10.0\% with increments of 1.0\%.  
\\

Lattice Configuration:\\
Here is the half-lattice map (symmetric arrangement assumed):  

\begin{verbatim}
1  
2  7  
3  8  5  
7  4  9  6  
4 10  5 11  5  
4  5 11  0  0  5  
7  5  6  0  0  5 10  
7  5  5  5  5  5  5  5  
3  6 10  5  5  5  5  5 10  
2  7  6  6  6  6  6  6  4  7
\end{verbatim}

Objective:  \\
Your goal is to generate new solutions that achieve:  \\
- Lattice criticality: \(kinf = 1.05 \)  \\
- Pin Peaking Factor (PPF): \( <1.30 \)  \\
- Higher score than any of the given solutions.  \\

The objective function (score) is determined based on these parameters, but you are not to guess \( k_{\inf} \), PPF, or score—these will be calculated externally using Casmo-5 simulations and a processing code.  \\

Solution Format:  \\
Each solution must include values for the following parameters:  \\
\texttt{FUE1\_enr, FUE2\_enr, FUE3\_enr, FUE4\_enr, FUE5\_enr, FUE6\_enr, FUE7\_enr, FUE8\_enr, FUE8\_gads, FUE9\_enr, FUE9\_gads, FUE10\_enr, FUE10\_gads, FUE11\_enr, FUE11\_gads}

Below are examples of solution-score pairs:

\begin{lstlisting}[basicstyle=\small\ttfamily\color{blue}]
<sol> 1.4,2.2,2.6,4.2,5.0,4.7,3.7,4.1,8.0,4.9,7.0,5.0,6.0,5.0,8.0 <\sol>
kinf: 1.03754, ppf: 1.361
score: 44.08

<sol> 1.8,2.4,2.9,4.0,5.0,4.7,3.4,3.9,5.0,4.6,5.0,5.0,8.0,5.0,7.0 <\sol>
kinf: 1.03530, ppf: 1.334
score: 66.6   

<sol> 1.8,2.5,2.7,4.2,5.0,4.9,3.7,3.8,8.0,4.7,5.0,5.0,8.0,4.8,9.0 <\sol>
kinf: 1.03643, ppf: 1.351
score: 51.86

\end{lstlisting}

Generate exactly \textcolor{red}{\{batch\_size\}} new solutions that:\\
- Are distinct from all previous solutions.\\
- Have a higher score than the highest provided.\\
- Respect the relationships between fuel enrichment, gadolinia content, kinf, and PPF based on logical reasoning.

The solutions should start with \texttt{<sol>} and end with \texttt{<\textbackslash sol>} with a comma between parameters. 

\end{tcolorbox}

The text in \textcolor{blue}{blue} is the solution-score pairs that are added every iteration.
\textcolor{red}{\{batch\_size\}} is to generate multiple solutions for every LLM response, currently set to 5 for this work.

\section{Samples of LLM Responses (Gemini 2.0 Flash)}
\label{app:responses}

\subsection{Responses to No Context Prompt}

\textbf{Step 0 Response (Best batch score: 51.92)}

\begin{tcolorbox}
\begin{verbatim}
<sol> 2.2,2.9,3.3,4.6,5.2,5.3,4.1,4.3,8.2,5.3,7.3,5.3,8.3,5.3,9.3 <\sol>
<sol> 2.3,3.0,3.4,4.7,5.3,5.4,4.2,4.4,8.3,5.4,7.4,5.4,8.4,5.4,9.4 <\sol>
<sol> 2.4,3.1,3.5,4.8,5.4,5.5,4.3,4.5,8.4,5.5,7.5,5.5,8.5,5.5,9.5 <\sol>
\end{verbatim}
\end{tcolorbox}

\textbf{Step 47 Response (Best batch score: 89.82)}

\begin{tcolorbox}
\begin{verbatim}
<sol> 1.8,2.5,3.0,4.3,6.0,5.0,3.8,4.0,8.1,5.0,7.1,5.1,8.1,5.1,13.8 <\sol>
<sol> 1.9,2.6,3.1,4.4,6.0,5.1,3.9,4.1,8.2,5.1,7.2,5.2,8.2,5.2,13.9 <\sol>
<sol> 2.0,2.7,3.2,4.5,6.0,5.2,4.0,4.2,8.3,5.2,7.3,5.3,8.3,5.3,14.0 <\sol
\end{verbatim}
\end{tcolorbox}

\subsection{Responses to Detailed Context Prompt}

\textbf{Step 0 Response (Best batch score: 81.16)}

\begin{tcolorbox}
Here are three new solutions, aiming for kinf closer to 1.05 and PPF below 1.30, while trying to be distinct from previous solutions. The reasoning is to increase enrichment in the central region and those pin locations which appear multiple times, and adjust gadolinia to control reactivity.

\begin{verbatim}
<sol> 1.9,2.6,2.8,4.3,5.0,4.8,3.6,4.0,7.0,4.8,6.0,5.0,7.0,4.9,8.0 <\sol>

<sol> 1.7,2.3,3.0,4.1,5.0,4.6,3.8,3.7,6.0,4.5,4.0,5.0,9.0,5.0,6.0 <\sol>

<sol> 2.0,2.7,2.9,4.4,5.0,4.9,3.5,4.2,9.0,4.9,8.0,5.0,5.0,5.0,7.0 <\sol>
\end{verbatim}

\end{tcolorbox}

\textbf{Step 9 Response (Best batch score: 91.36)}

\begin{tcolorbox}
Based on the provided solutions, I observe the following trends:\\

- Higher enrichments (generally) lead to higher kinf.\\
- Introducing Gadolinia generally decreases kinf, but can help with PPF.\\
- There's no clear single enrichment or gadolinia that dominates in terms of score. It's a balance.\\
- Solutions with kinf significantly above 1.05 tend to not be scored highly, likely due to the PPF penalty as criticality goes up.\\
- Lower PPF generally improves score, while values above 1.4 lead to severe score reduction.\\

To generate new solutions with a higher score than the existing maximum (82.42), I need to find a configuration that maximizes kinf without exceeding 1.05 too much, and keeps PPF low. I will focus on solutions where kinf is close to 1.05.  I'll prioritize adjustments to fuel types that seem influential (like FUE4, FUE5, and FUE9, based on their frequency of change in top-scoring examples). I will also try adjustments to gadolinia content.\\

Here are three new solutions:\\
\begin{verbatim}
<sol> 2.1,2.7,2.9,4.3,5.0,4.8,3.6,4.0,8.0,4.8,7.0,5.0,5.0,4.9,6.0 <\sol>
<sol> 2.2,2.8,3.0,4.5,5.0,4.9,3.8,4.1,6.0,4.9,5.0,5.0,8.0,4.7,7.0 <\sol>
<sol> 2.0,2.6,3.1,4.4,5.0,4.7,3.9,4.2,7.0,4.9,6.0,5.0,6.0,5.0,8.0 <\sol>
\end{verbatim}

\end{tcolorbox}

\end{document}